\documentclass[10pt,twocolumn,letterpaper]{article}

\usepackage[pagenumbers]{cvpr} % To force page numbers, e.g. for an arXiv version

\usepackage{amsmath,amsfonts,bm}

\def\eqref#1{equation~\ref{#1}}
\def\ceil#1{\lceil #1 \rceil}

\def\1{\bm{1}}

\DeclareMathAlphabet{\mathsfit}{\encodingdefault}{\sfdefault}{m}{sl}
\SetMathAlphabet{\mathsfit}{bold}{\encodingdefault}{\sfdefault}{bx}{n}

\def\gN{{\mathcal{N}}}

\usepackage{algorithm}% http://ctan.org/pkg/algorithms
\usepackage{algpseudocode}% http://ctan.org/pkg/algorithmicx
\usepackage{amsmath}
\usepackage{multirow}
\usepackage{xspace}
\usepackage{makecell}
\newcommand{\boldtheta}{\boldsymbol{\theta}}

\definecolor{cvprblue}{rgb}{0.21,0.49,0.74}
\usepackage[pagebackref,breaklinks,colorlinks,allcolors=cvprblue]{hyperref}

\makeatletter
\DeclareRobustCommand\onedot{\futurelet\@let@token\@onedot}
\def\@onedot{\ifx\@let@token.\else.\null\fi\xspace}

\def\ie{\emph{i.e}\onedot}

\makeatother

\newcommand{\shortmodelname}{QDM\xspace}

\def\paperID{10443} % *** Enter the Paper ID here
\def\confName{CVPR}
\def\confYear{2026}

\title{QDM: Quadtree-Based Region-Adaptive Sparse Diffusion Models \\ for Efficient Image Super-Resolution}

\author{Donglin Yang\\
The University of Hong Kong\\
{\tt\small ydlin0718@connect.hku.hk}
\and
Paul Vicol\\
Google DeepMind\\
{\tt\small paulvicol@google.com}
\and
Xiaojuan Qi\\
The University of Hong Kong\\
{\tt\small xjqi@eee.hku.hk}
\and
Renjie Liao\\
The University of British Columbia\\
{\tt\small renjie.liao@ubc.ca}
\and
Xiaofan Zhang\\
Shanghai Jiao Tong University\\
{\tt\small xiaofan.zhang@sjtu.edu.cn}
}

\begin{document}
\maketitle

\begin{abstract}

Deep learning-based super-resolution (SR) methods often perform pixel-wise computations uniformly across entire images, even in homogeneous regions where high-resolution refinement is redundant. 
We propose the Quadtree Diffusion Model (QDM), a region-adaptive diffusion framework that leverages a quadtree structure to selectively enhance detail-rich regions while reducing computations in homogeneous areas.
By guiding the diffusion with a quadtree derived from the low-quality input, QDM identifies key regions—represented by leaf nodes—where fine detail is essential and applies minimal refinement elsewhere.
This mask-guided, two-stream architecture adaptively balances quality and efficiency, producing high-fidelity outputs with low computational redundancy.
Experiments demonstrate QDM’s effectiveness in high-resolution SR tasks across diverse image types, particularly in medical imaging (e.g., CT scans), where large homogeneous regions are prevalent.
Furthermore, QDM outperforms or is comparable to state-of-the-art SR methods on standard benchmarks while significantly reducing computational costs, highlighting its efficiency and suitability for resource-limited environments. Our code is available at \url{https://github.com/linYDTHU/QDM}.
\end{abstract}

\section{Introduction}
\label{sec:intro}
Super-resolution (SR) seeks to reconstruct high-resolution (HR) images from low-resolution (LR) inputs, an inherently ill-posed inverse problem due to the unknown and complex image formation processes in real-world scenarios. 
Existing approaches mitigate this challenge by designing empirical image formation pipelines and leveraging generative adversarial networks (GANs)~\cite{goodfellow2020generative} to approximate the LR-to-HR mapping~\cite{zhang2021designing,wang2021real,liang2021swinir}. 
Recent advances in diffusion models~\cite{ho2020denoising,sohl2015deep,song2020score} have demonstrated superior image generation capabilities compared to GANs~\cite{dhariwal2021diffusion}, positioning them as promising tools for SR. 
Contemporary studies achieve this either by adapting diffusion frameworks for SR tasks~\cite{saharia2022image,rombach2022high,yue2024resshift,wang2024sinsr} or exploiting pretrained text-to-image (T2I) models like Stable Diffusion~\cite{rombach2022high,podell2023sdxl} as priors to regularize the ill-posed reconstruction~\cite{lin2024diffbir,wu2024seesr,wang2024exploiting,wu2025one, yue2024arbitrary, yang2024pixel}.

\begin{figure}[t]  % Use [t] to place it at the top of the column; other options are [b] (bottom) and [h] (here)
    \centering
    \includegraphics[width=\columnwidth]{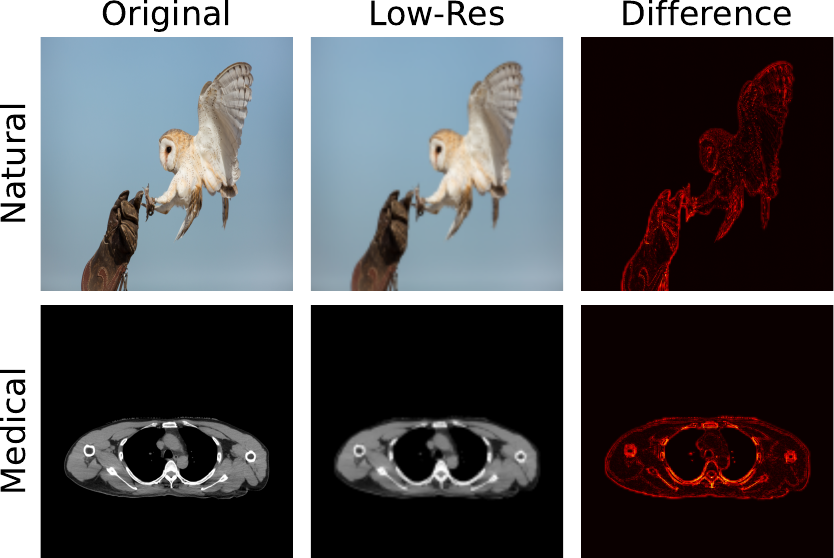}  % \columnwidth scales it to fit one column
    % \vspace{-0.6cm}
    \caption{The original high-resolution (HR) natural images collected from DIV2K \cite{agustsson2017ntire} and medical images from SegRap2023 \cite{luo2025segrap2023}, are downsampled to lower resolutions. The difference heatmaps reveal that only sparse detail-rich regions—such as edges and textures—require significant refinement, while large homogeneous areas remain largely unaffected.}
    % \vspace{-0.6cm}
    \label{fig:intuition demonstration}
\end{figure}

Despite recent advances, most SR methods apply uniform pixel-wise computations across the entire image, regardless of local content variations. 
While this type of approach effectively enhances details, it incurs unnecessary computational overhead, especially in homogeneous regions where pixel intensities are similar, \ie, exhibiting redundant information.  
As illustrated in Fig.~\ref{fig:intuition demonstration}, fine-grained details benefit the most from the SR process, whereas homogeneous regions require minimal processing. 
However, existing SR methods largely overlook this spatial adaptivity, leading to redundant computations.

To address these inefficiencies, we propose an adaptive computational framework that selectively allocates computations to detail-rich regions while minimizing computation in homogeneous areas. 
Our approach resolves two key challenges: \textit{ 1) identifying homogeneous regions that require minimal processing while preserving structures in detail-rich areas, and 2) designing an architecture capable of spatially adaptive computation}. 

For the first challenge, we employ a quadtree decomposition strategy~\cite{samet1984quadtree}.
This approach recursively subdivides regions until they reach a predefined homogeneity threshold, resulting in variable-sized homogeneous regions at the leaf nodes.
We can then leverage this quadtree prior to refining regions only where necessary, thus reducing redundant computations.
For the second challenge, existing diffusion-based SR methods~\cite{ho2020denoising,song2020score,rombach2022high,yue2024resshift} typically rely on architectures, such as U-Net~\cite{ronneberger2015u} or Transformer-based models~\cite{peebles2023scalable}, which uniformly process all spatial regions without adaptivity. 
We instead propose a dual-stream architecture guided by the quadtree mask. 
Our design processes coarse-grained patches in upstream blocks to capture global context, while dynamically routing detail-rich regions to downstream blocks for enhancing fine-grained patches. 
Our approach provides two main advantages: (1) it localizes computationally intensive operations to detail-rich regions, and (2) it partitions selected tokens into parallelizable chunks, eliminating the need for full-feature-map storage.

In summary, the main contributions of this work are twofold. First, we propose a hierarchical quadtree-based algorithm to efficiently identify detail-rich regions, and design mask-guided diffusion processes that selectively refine these regions while leveraging homogeneous areas as spatial-contextual priors. 
Second, we introduce a dual-stream architecture with dynamic resource allocation, enabling spatial-adaptive computation to improve processing efficiency.
Extensive experimental results show that our \shortmodelname outperforms or matches state-of-the-art SR methods on both natural and medical imaging SR benchmarks, while reducing computational costs.

\section{Related Works}
\label{sec:related works}

\noindent \textbf{Diffusion-based Image Super-Resolution.} 
Diffusion-based models for super-resolution (SR)  have been widely explored and can be roughly categorized into two approaches: train-from-scratch and pretrained-prior-based methods. 
Train-from-scratch techniques, initiated by SR3~\cite{saharia2022image}, pioneered the application of diffusion models to SR. 
LDM~\cite{rombach2022high} further enhanced computational efficiency by operating in a latent space, while ResShift~\cite{yue2024resshift} further streamlined the process to $15$ diffusion steps via residue shifting. 
These methods are easy to retrain, and inherently versatile, making them broadly applicable across diverse scenarios.
Another research direction leverages the generative power of pretrained text-to-image (T2I) models like Stable Diffusion~\cite{rombach2022high, podell2023sdxl}. 
Works like DiffBIR~\cite{lin2024diffbir}, SeeSR~\cite{wu2024seesr}, and StableSR~\cite{wang2024exploiting}, along with others~\cite{wu2025one, yue2024arbitrary, yang2024pixel}, employ various fine-tuning strategies to exploit diffusion priors for SR, achieving state-of-the-art performance. 
However, such methods rely heavily on large-scale T2I foundation models, limiting their adaptability to domains beyond natural images.
Our approach falls within the train-from-scratch paradigm, prioritizing flexibility and broad applicability without using external pretrained models.

\noindent \textbf{Model Architecture for Diffusion Models.}
The rapid advancement of diffusion models has established U-Net architectures~\cite{ronneberger2015u} as foundational to diffusion-based image synthesis. 
These architectures have demonstrated exceptional performance across tasks spanning denoising, super-resolution, and high-definition image generation~\cite{ho2020denoising, song2020score, rombach2022high, yue2024resshift}. 
Early frameworks like DDPMs \cite{ho2020denoising} augmented U-Nets with self-attention layers at coarse resolutions to enhance feature interaction. 
Later studies \cite{hoogeboom2023simple, podell2023sdxl} emphasize the critical role of additional self-attention in low-resolution stages for scaling to high-dimensional outputs. 
Concurrently, the Swin Transformer \cite{liu2021swin} introduces shifted-window self-attention, which confines attention to localized windows while preserving global context through window shifting. 
This approach has been successfully integrated into diffusion frameworks like ResShift \cite{yue2024resshift} and SinSR \cite{wang2024sinsr} to balance scalability and efficiency. 
Recently, DiT \cite{peebles2023scalable} has emerged as a milestone, replacing U-Nets entirely with pure Transformer backbones, showcasing unprecedented scalability for large-scale generative tasks.

\noindent \textbf{Quadtree Representations.}
Quadtrees are hierarchical data structures with a long history in image processing \cite{samet1984quadtree}. 
Early works, such as \cite{bouman1994multiscale, feng1998training}, applied quadtrees to semantic segmentation tasks using graphical models. 
However, in the deep learning era, their use has been primarily focused on reducing computation and memory costs. 
For instance, \cite{jayaraman2018quadtree} presents a quadtree-based convolutional network for tasks involving sparse, binary input images. 
Similarly, \cite{chitta2020quadtree} leverages an efficient quadtree representation of segmentation masks to reduce memory consumption with minimal accuracy loss. 
Quadtree attention \cite{tang2022quadtree} computes attention in a coarse-to-fine manner, reducing computational complexity from quadratic to linear.

\section{Method}
\label{sec:method}

In this section, we first describe how to generate quadtree masks from a given LR image, enabling the identification of detail-rich regions for super-resolution.
Next, we explain how these masks guide the diffusion process to focus computation on these important areas.
Finally, we introduce our dual-stream model architecture, designed for adaptive computation and improved efficiency.

\subsection{Quadtree Prior---Identify Detail-Rich Regions}
\label{subsec:quadtrees masks}

A quadtree is a hierarchical data structure where each non-leaf node has four child nodes \cite{samet1984quadtree}. 
This structure recursively partitions a two-dimensional space into four quadrants or regions. 
If all pixels within a sub-region exhibit similar values (\ie, low variance), the sub-region becomes a leaf node in the quadtree and is represented by a single aggregated value. 
In this work, we adopt a lossy compression strategy, considering a sub-region as a leaf node if the $l_1$ norm of the difference between its maximum and minimum values is below a threshold $s$. 
This approach maximizes compression while preserving essential image details.

Using this scheme, we construct a hierarchical quadtree structure from a low-resolution (LR) image $Y_0$. 
The bottom level of the quadtree represents the finest granularity, corresponding to the resolution of the LR image. 
A key assumption in our approach is that \textit{LR and HR images share the same structural hierarchy at upper levels}. 
Consequently, leaf nodes at higher levels require minimal refinement, as they represent relatively homogeneous regions. 
In contrast, bottom-level nodes may require further subdivision and refinement to capture finer details. 
Thus, the primary objective is to efficiently identify the positions of these bottom-level leaf nodes.
We propose to introduce a binary mask $M$ to store leaf node positions efficiently.
Here, ``1''s indicate the presence of leaf nodes at the bottom level, while ``0''s correspond to homogeneous regions.
These binary masks are efficiently generated using Algorithm \ref{alg:quadtree masks generation}, which is optimized for parallel execution on GPUs. 
Our method constructs each binary mask with minimal computational overhead, making it well-suited for real-time applications and large-scale data processing. 
Crucially, it's important to note that the generated mask can have the same spatial resolution with latent feature maps rather than requiring alignment with the input LR image, enabling seamless compatibility with latent diffusion models~\cite{rombach2022high}. 

\begin{algorithm}[t]
    \caption{Efficient Generation of Binary Quadtree Masks using Max Pooling}
    \label{alg:quadtree masks generation}
    \begin{algorithmic}[1]
        \State \textbf{Input:} A batch of low-quality images $I\in \mathbb{R}^{B\times C\times H\times W}$, threshold $s$, desired mask height $H_m$ and desired mask width $W_m$
        \State \textbf{Output:} Binary Quadtree Masks $M \in \left\{0, 1\right\}^{B\times H_m \times W_m}$
        \State $l=\max\left(\log_{2}{\frac{H}{H_m}}, \log_{2}{\frac{W}{W_m}}\right)$
        \State $l=\max\left(\ceil{l}, 1\right)$

        %Generate mask for level $l$
        \State $V^{+} \leftarrow$ Max pooling on $I$ with kernel size $2^l \times 2^l$
        \State $V^{-} \leftarrow$ Max pooling on $-I$ with kernel size $2^l \times 2^l$
        \State $M \leftarrow \left[\left(V^{+} + V^{-}\right)\geq s\right]$
        \State $M \leftarrow$ Nearest-neighbor interpolation on $M$ to the size $(H_m, W_m)$
        \State \textbf{return} $M$
    \end{algorithmic}
\end{algorithm}

\subsection{Mask-Guided Sparse Diffusion Process}
\label{subsec:mask-guided diffusion process}

Given a LR image $Y_0$, our goal is to generate the corresponding HR image $X_0$. 
Following standard notation in the diffusion literature, the subscript denotes the diffusion time step, where $0$ corresponds to data and $1$ to pure noise. 
As discussed in Section~\ref{subsec:quadtrees masks}, refining large homogeneous regions in an LR image is often unnecessary. 
To address this, we propose a mask-guided diffusion process, where the mask, often being sparse, is derived from the quadtree representation of the LR image. 
This approach reduces computational redundancy and reallocates model capacity toward refining detail-rich regions. 
In the following, we introduce the forward and reverse processes of \shortmodelname, along with the corresponding training objective.

\begin{figure*}[ht]  % Use [t] to place it at the top of the column; other options are [b] (bottom) and [h] (here)
    \centering
    \includegraphics[width=\linewidth]{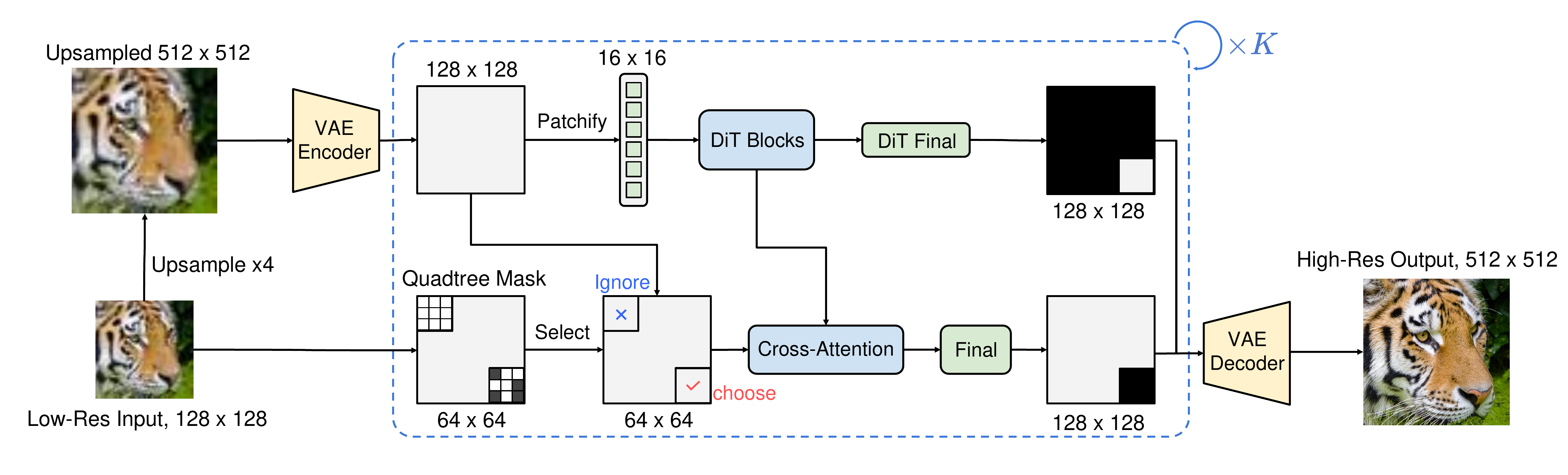}
    % \vspace{-0.6cm}
        \caption{\textbf{Dual-Stream Architecture Overview.} Our transformer-based model employs two complementary processing streams guided by a quadtree mask for input-adaptive computation. The upstream (top) extracts global context via Diffusion Transformer (DiT) blocks operating on patches with size 8$\times$8, while the downstream (bottom) refines patches with size 2$\times$2 exclusively in mask-selected regions. Mask $M$ dynamically controls computational overhead by routing computationally intensive processing only to detail-rich areas. Selected tokens are partitioned into parallelizable chunks, each processed independently through cross-attention blocks. Final predictions combine the coarse upstream outputs with the refined downstream results. }
    \label{fig:framework}
    \vspace{-0.5cm}
\end{figure*}

\begin{figure*}[ht]  % Use [t] to place it at the top of the column; other options are [b] (bottom) and [h] (here)
    \centering
    \includegraphics[width=0.9\linewidth]{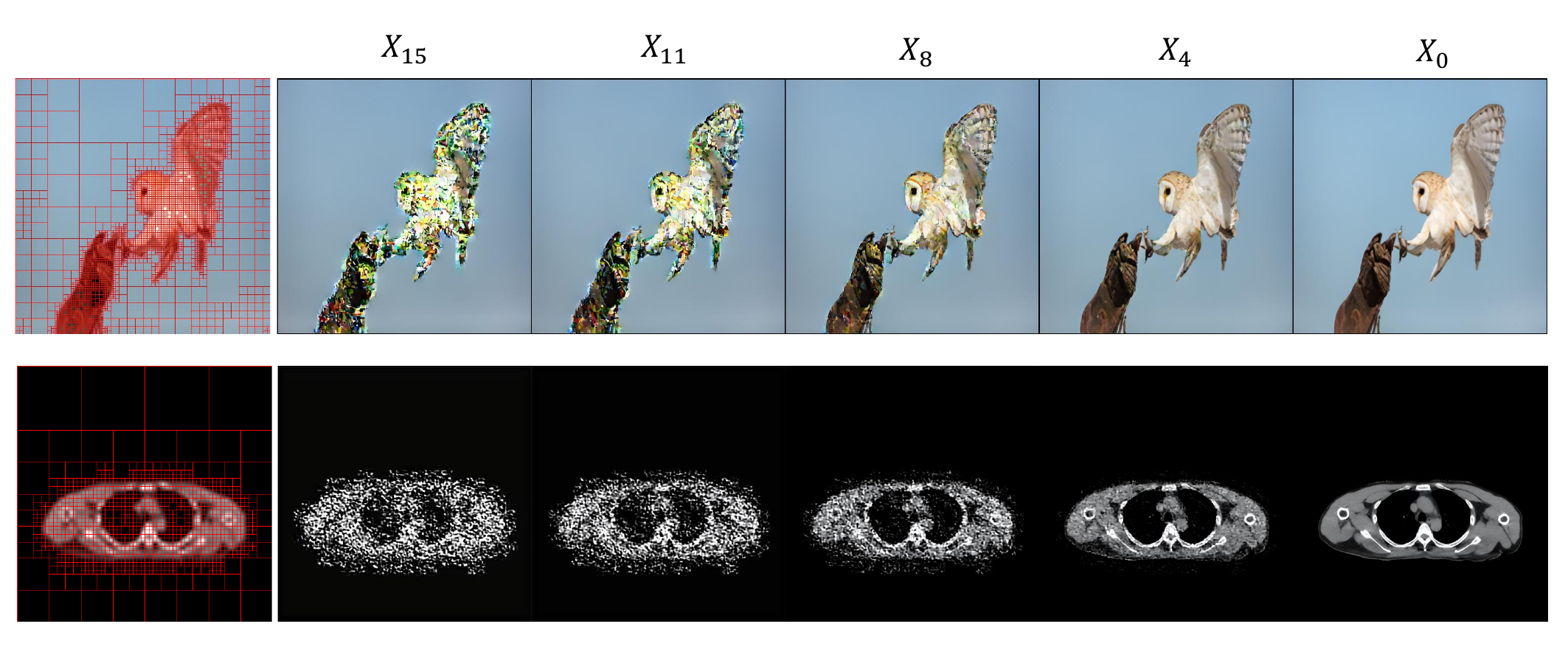}
    % \vspace{-0.25cm}
    \caption{\textbf{Examples of Mask-Guided Diffusion Process.} Left: Low-quality inputs with quadtree partitions (thresholds $s=0.06$ for top, $s=0.00$ for bottom). Right: Sampling trajectories of our QDM framework. Our diffusion process selectively adds noise to detail-rich regions (quadtree leaf nodes, smallest blocks in the left images that are often sparse) while preserving homogeneous areas.}
    \vspace{-0.3cm}
    \label{fig:diffusion process}
\end{figure*}

\noindent \textbf{Forward Process.} 
In the forward process, we gradually add noise to the HR image $Y_0$ conditioned on the LR image $X_0$.
In contrast to vanilla diffusion models where the forward process adds noise everywhere, our approach only adds noise at active sites, \ie, detail-rich regions where the redundancy is low. 
In particular, we first obtain a mask $M$ of active sites via Algorithm \ref{alg:quadtree masks generation}, where entries of $M$ with value 1 indicate the active sites. 
Inspired by the work \cite{yue2024resshift}, we design the following transition distribution of the forward process that focuses solely on these active sites,
\begin{align}
\label{eq:transition kernel}
& q(X_t \mid X_{t-1}, Y_0, M) =\gN \left(X_t; \bm{\mu}_t, \mathrm{\text{diag}}(\bm{\sigma}_t)\right), \nonumber \\
& \bm{\mu}_t = (\alpha_t E_0 + X_{t-1})\odot M  + (\eta_t E_0 + X_0) \odot (1-M) \nonumber \\
& \bm{\sigma}_t = \kappa^2\alpha_t M, 
\end{align}
where $E_0 = Y_0 - X_0$ is the residual error between the LR and the HR images and $\odot$ denotes the element-wise product.
Thus, during the forward process, we progressively transform the HR image $X_0$ to the LR image $Y_0$, selectively applying noise to active regions while preserving non-active areas.
The shifting sequence $\{\eta_t\}_{t=1}^T$ is a monotonically increasing sequence that satisfies $\eta_1 \rightarrow 0$ and $\eta_T \rightarrow 1$, $\alpha_t = \eta_t - \eta_{t-1}$ for $t>0$ and $\kappa$ controls the noise magnitude.  
We can derive the following conditional distribution,
\begin{align}\label{eq:marginal distribution}
q(X_t \mid X_0, & Y_0, M) =  \gN \left(X_t; \eta_t E_0 + X_0 , \mathrm{\text{diag}}(\kappa^2\eta_t M)\right).
\end{align}
\noindent \textbf{Reverse Process.}
In the reverse process, we learn to perform denoising to recover the HR image. 
In particular, we have the following posterior distribution: 
\begin{align}\label{eq:posterior_distribution}
    p_{\bm{\theta}} \left(X_0 \mid Y_0, M\right) = 
    &\int p(X_T \mid Y_0, M) \prod_{t=1}^T \nonumber \\
    & p_{\bm{\theta}} \left(X_{t-1} \mid X_t, Y_0, M\right) \,\text{d}X_{1:T},
\end{align}
where $p(X_T \mid Y_0, M) = \gN \left(X_T; Y_0, \mathrm{\text{diag}}(\kappa^2 M) \right)$ and $p_{\bm{\theta}} \left(X_{t-1} \mid X_t, Y_0, M\right)$ is the reverse transition probability. 
Following common practice in diffusion models \cite{sohl2015deep, ho2020denoising, song2020score, yue2024resshift}, we parameterize the reverse transition probability as:
\begin{align}
\label{eq:paramterization}
    & p_{\bm{\theta}} \left(X_{t-1} \mid X_t, Y_0, M\right) = \gN \left(X_{t-1}; \hat{\bm{\mu}}, \mathrm{\text{diag}}(\kappa^2\frac{\eta_{t-1}}{\eta_{t}}\alpha_t M) \right) \nonumber \\
    & \hat{\bm{\mu}} = \frac{\eta_{t-1}}{\eta_t} \odot X_t + \frac{\alpha_t}{\eta_t}  \odot f_{\bm{\theta}}(X_t, Y_0,t), 
\end{align}
where $f_{\boldtheta}$ is a deep neural network with parameters $\boldtheta$, tailored to accurately estimate pixel values on active sites while making only minimal adjustments to the remaining regions. During the reverse process, the model iteratively denoises the active (detail-rich) regions, while applying minimal deterministic refinement to non-active (homogeneous) areas to preserve their structures. This procedure is illustrated in Figure~\ref{fig:diffusion process}.
The training objective is to minimize the following mean squared error,
\begin{align}
\label{eq: training objective function}
    \min_{\boldtheta} \quad \mathcal{L}_{\boldtheta} =  \mathbb{E}_{X_0, Y_0, t}   \left \Vert f_{\boldtheta}(X_t, Y_0,t) - X_0 \right \Vert_2^2 .
\end{align}
We also adopted the non-uniform geometric schedule for the diffusion time $t$ as proposed in \cite{yue2024resshift}.

\noindent \textbf{Extension to Latent Space.}
To reduce computational costs, recent studies \cite{rombach2022high, peebles2023scalable, yue2024resshift} propose training diffusion models in the latent space of a pretrained variational autoencoder (VAE). 
In our framework, the encoder compresses images, reducing their spatial dimensions, which introduces a size mismatch between the compressed latent maps and the low-quality images. 
Fortunately, Alg.~\ref{alg:quadtree masks generation} can generate masks that align with the resolution of the latent maps. 
The spatial locations of active regions in the original image are preserved in the latent space due to the translation-equivariant properties of the convolutional encoder. 
Adapting our model to the latent space only requires replacing $X_0$ and $Y_0$ with their latent representations, along with the appropriately sized mask $M$.

\subsection{Model Architecture}
\label{subsec:model architecture}

Fully transformer-based architectures \cite{peebles2023scalable} have outperformed the conventional U-Net models \cite{ronneberger2015u} in many tasks like image super-resolution. 
However, existing approaches are limited by their computationally expensive full-sequence attention mechanisms.
This leads to substantial memory consumption and inefficient resource allocation, particularly in super-resolution tasks where only certain detail-rich regions necessitate complex processing, while others could be handled with simpler transformations.

To address these limitations, we propose a novel dual-stream transformer architecture that integrates three key innovations: (1) region-adaptive computation guided by quadtree-based spatial partitioning, (2) dynamic memory allocation through hierarchical feature processing, and (3) context-aware token independence with global coordination. As illustrated in Figure \ref{fig:framework}, our framework employs parallel processing streams - an upper stream preserving global contextual relationships and a lower stream specializing in localized feature refinement. 
This bifurcated approach enables selective resource allocation, directing complex transformations toward detail-rich regions identified by our adaptive masking mechanism while maintaining spatial coherence through integrated prediction.

\begin{figure*}[ht]
\centering
\includegraphics[width=0.9\textwidth]{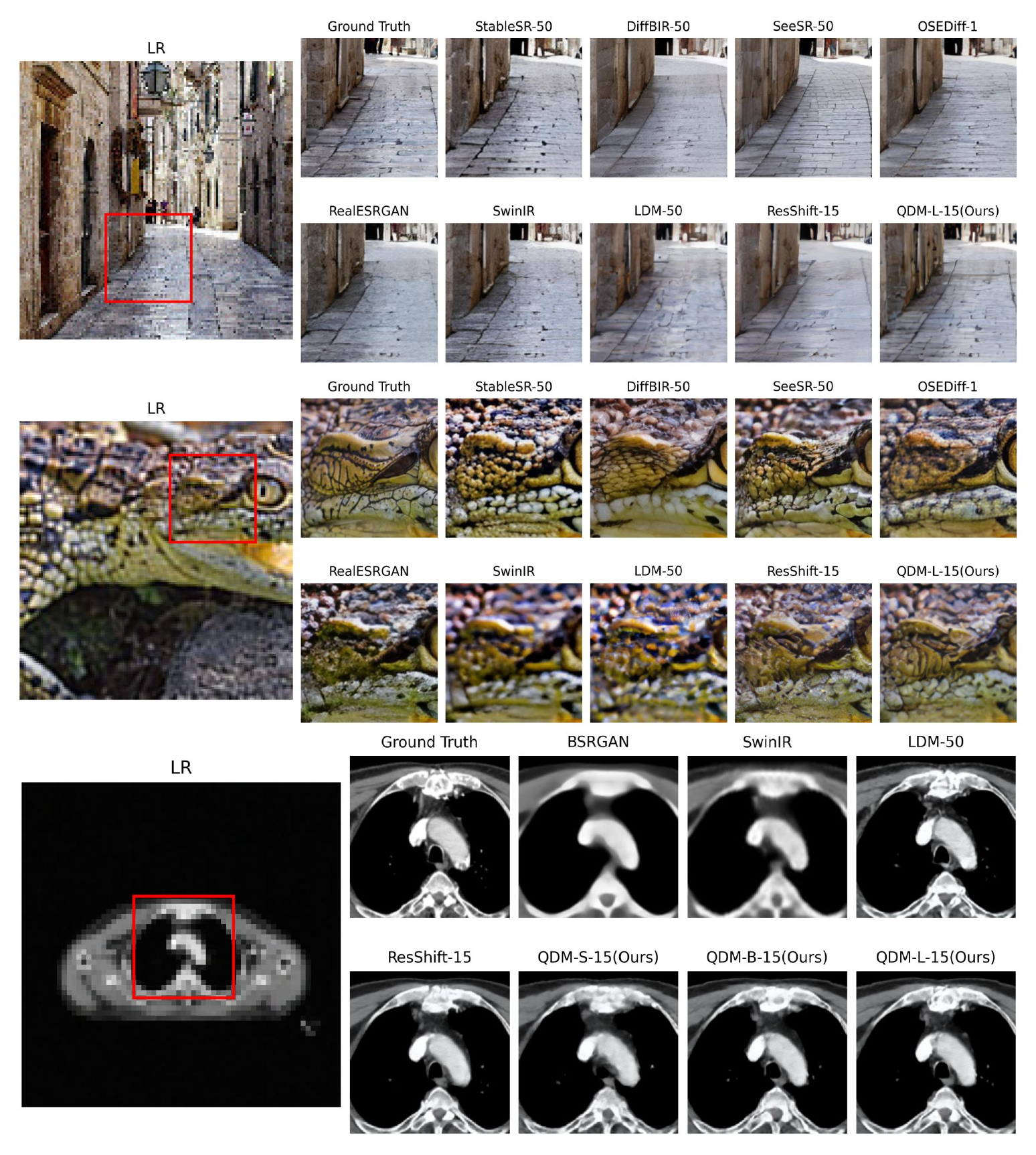}
\caption{Visual comparison of different methods on real-world images and medical CT datasets. Zoom in for finer details. }  
\label{fig:comparison}
\end{figure*}

\noindent \textbf{Upstream.}
The upstream branch utilizes a Diffusion Transformer (DiT)~\cite{peebles2023scalable} with a large patch size of 8 to extract global features from the input image. This design enables coarse refinements in homogeneous regions while reducing the token count by $64\times$ compared to pixel-level processing. By operating on coarse-grained patches, the model significantly lowers memory consumption and inference time while preserving global context.
The large patch size expands the receptive field, improving efficiency in high-level semantic feature extraction. These global features are then propagated to downstream modules via cross-attention layers, enhancing spatial consistency in predictions.

\noindent \textbf{Downstream.}  
The downstream branch works in tandem with the upstream one, refining key regions identified by the quadtree mask $M$ (Alg.~\ref{alg:quadtree masks generation}). Given an input of size $H \times W \times C$, the image is first divided into $2 \times 2$ patches, forming a feature grid of dimensions $\frac{H}{2} \times \frac{W}{2} \times C$. Following the Swin Transformer~\cite{liu2021swin}, this grid is further partitioned into non-overlapping $N \times N$ local windows, producing $\frac{HW}{4N^2}$ feature chunks of size $N^2 \times C$. The quadtree mask $M$ (size $\frac{H}{2} \times \frac{W}{2} \times C$) selectively determines which chunks require computation: only windows overlapping with active sites (mask entries being $1$) are projected into a high-dimensional space. These selected features are then augmented with 2D positional embeddings to preserve spatial relationships.  

Each feature chunk is processed through a series of attention-based operations. It first undergoes multi-head self-attention (MSA) to model local dependencies, followed by multi-head cross-attention (MCA) to incorporate global context from the upstream branch's coarse features (dimension $C_{\text{up}}$, size $\frac{HW}{64} \times C_{\text{up}}$). The output is then passed through a two-layer feedforward network with GELU activation and time-conditioned via adaLN-Zero layers~\cite{peebles2023scalable}. Residual connections are applied after each MSA, MCA, and feedforward operation.  
To improve efficiency, the upstream’s global features are pre-projected into key-value pairs ($K, V$) and cached for reuse across all chunks during inference, minimizing redundant computations. The final projection restores features to their original dimensionality. 

\begin{figure}[t]
    \centering
    % First subfigure: Inference Speed Comparison
    \includegraphics[width=\columnwidth]{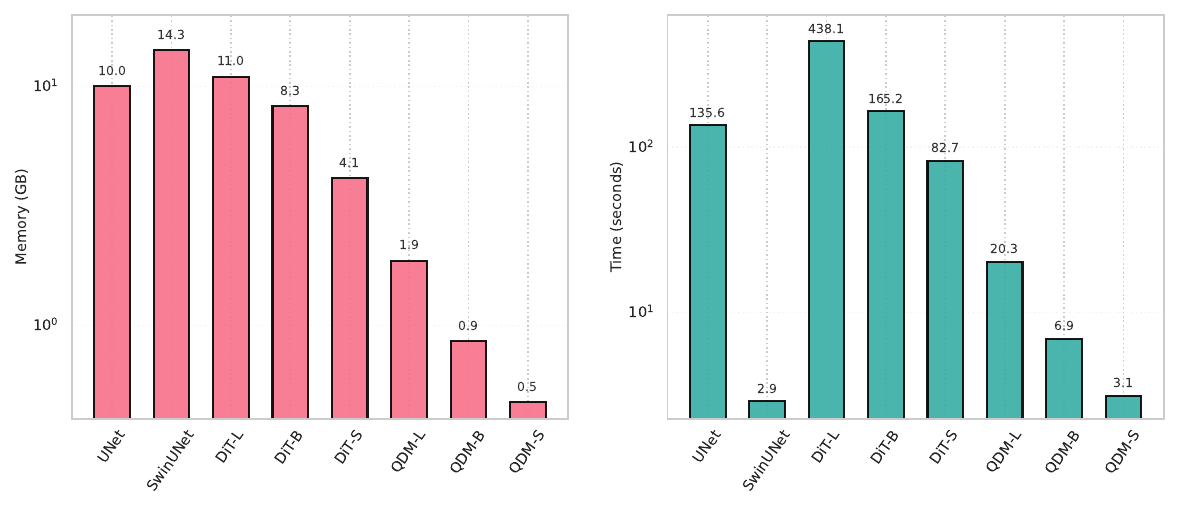}
% \vspace{-0.5cm}
    \caption{\textbf{Comparison of QDM, DiT~\cite{peebles2023scalable}, U-Net~\cite{ronneberger2015u}, and a Swin Attention-enhanced U-Net variant~\cite{liu2021swin} with high-resolution ($1024\times1024$) inputs.} The left panel shows peak memory usage, while the right panel depicts inference speed across evaluated models. All metrics were measured on an NVIDIA A100 80G GPU. For QDM, the evaluation was conducted under the most resource-intensive condition, using a full mask without adaptive computation and processing 64 chunks in parallel.}
    \vspace{-0.5cm}
    \label{fig:efficiency-comparison}
\end{figure}

\noindent \textbf{Dual-Stream Inference.}  
The dual-stream architecture adaptively allocates computation based on input complexity. Homogeneous regions are efficiently processed by the upstream branch, while the downstream branch focuses on detail-rich areas identified by the quadtree mask $M$. The mask's density threshold $s$ in Alg.~\ref{alg:quadtree masks generation} controls this allocation: higher $s$ values increase the sparsity of the mask, reducing FLOPs in uniform regions. 
By dynamically routing computations to finer $2 \times 2$ patches only where needed, the model achieves input-dependent efficiency while ensuring high-quality refinements.  
We also conduct a comprehensive comparison of our model against state-of-the-art architectures, evaluating both computational efficiency and peak memory utilization on high-resolution inputs. As illustrated in Figure~\ref{fig:efficiency-comparison}, our proposed QDM framework demonstrates substantial improvements across both metrics. Specifically, QDM achieves a significant reduction in memory consumption while simultaneously accelerating inference speed, outperforming existing approaches in both aspects. This dual advantage highlights the scalability and practical applicability of QDM for resource-intensive tasks.

\noindent \textbf{Training.}  
The training objective combines dual-stream predictions through a weighted loss:
\begin{align}
    \min_{\bm{\theta}} ~~\lambda_1 \mathcal{L}_{\text{down}} + \lambda_2 \mathcal{L}_{\text{up}},
    \label{eq:dual-stream training objective}
\end{align}
where $\mathcal{L}_{\text{up}}$ and $\mathcal{L}_{\text{down}}$ denote upstream and downstream prediction losses that follow the form in Eq.~(\ref{eq: training objective function}), and $\lambda_1,\lambda_2$ control the relative importance.
To enhance photorealism, we incorporate the perceptual loss:
\begin{align}
    \min_{\bm{\theta}} ~~ \lambda_1 \mathcal{L}_{\text{down}} + \lambda_2 \mathcal{L}_{\text{up}} + \lambda_3 \mathcal{L}_{\text{LPIPS}}(f_{\bm{\theta}}(X_t,Y_0, t), X_0),
    \label{eq:perceptual training objective}
\end{align}
where $\mathcal{L}_{\text{LPIPS}}$~\cite{zhang2018unreasonable} measures perceptual similarity between prediction $f_{\bm{\theta}}(X_t,Y_0,t)$ and ground-truth $X_0$, with $\lambda_3$ balancing detail preservation against prediction accuracy.

\noindent \textbf{Model Size.} We apply a sequence of $N$ attention blocks for both upstream and downstream. Following DiT~\cite{peebles2023scalable}, we use standard transformer configurations that jointly scale up the number of attention blocks, the hidden size, and the number of attention heads. 
Specifically, we use three configs, QDM-S, QDM-B and QDM-L, aligned in size with the DiT variants. Appendix \ref{appendix:configurations} provides more details of these configurations.

\section{Experiments}
\label{sec:exp}

\begin{table*}[!htbp]
\centering
\caption{Quantitative comparisons on the \textit{LSDIR-Test}, \textit{RealSR}, and \textit{RealSet80} datasets. The best and second-best results within each paradigm are highlighted in \textbf{bold} and \underline{underlined}, respectively. The number following each method name (e.g., -15 or -50) indicates the number of inference steps used. All results are averaged over three runs with different random seeds.}
\vspace{-0.2cm}
\resizebox{0.9\textwidth}{!}{
\begin{tabular}{lccccccccc}
\toprule
\multirow{2}{*}{\textbf{Method}} 
& \multicolumn{5}{c}{\textbf{LSDIR-Test}} 
& \multicolumn{2}{c}{\textbf{RealSR}} 
& \multicolumn{2}{c}{\textbf{RealSet80}} \\
\cmidrule(lr){2-6} \cmidrule(lr){7-8} \cmidrule(lr){9-10}
& PSNR$\uparrow$ & SSIM$\uparrow$ & LPIPS$\downarrow$ & CLIPIQA$\uparrow$ & MUSIQ$\uparrow$
& CLIPIQA$\uparrow$ & MUSIQ$\uparrow$
& CLIPIQA$\uparrow$ & MUSIQ$\uparrow$ \\
\midrule
\multicolumn{10}{l}{\textit{Pretrained-Prior-Based Methods}} \\
StableSR-50~\cite{wang2024exploiting}   
&20.40 &0.5448 &0.2630 &0.6821 &68.6041  
&0.5209 &60.1758  
&0.6214 &62.7613 \\
DiffBIR-50~\cite{lin2024diffbir}   
&21.28 &0.5222 &0.2813 &0.7406 &72.8067  
&0.6857 &65.3934  
&\textbf{0.7404} &67.9806 \\
SeeSR-50~\cite{wu2024seesr}   
&\underline{21.36} &0.5506 &\underline{0.2686} &\underline{0.7223} &\textbf{73.2258}  
&0.6824 &\underline{66.3757}  
&0.7114 &\underline{69.7658} \\
OSEDiff-1~\cite{wu2025one}   
&21.17 &\underline{0.5517} &0.2723 &0.6896 &72.1574  
&\textbf{0.7008} &65.4806  
&0.7093 &68.8022 \\
InvSR-1~\cite{yue2024arbitrary}  
&20.23 &0.5413 &0.2747  &0.7213 &\underline{73.1948}  
&\underline{0.6918} &\textbf{67.4586}  
&\underline{0.7291} &\textbf{69.8055} \\
SinSR-1~\cite{wang2024sinsr}   
&\textbf{21.99} &\textbf{0.5750} &\textbf{0.2520} &0.6683 &69.2179  
&0.6634 &59.2981  
&0.7228 &64.0573 \\
\midrule
\multicolumn{10}{l}{\textit{Training-from-Scratch Methods}} \\
BSRGAN~\cite{zhang2021designing}  
&22.05 &0.5795 &0.2929 &0.5898 &\underline{69.1739}
&0.5439 &\textbf{63.5869}  
&0.6263 &\textbf{66.6288} \\
RealESRGAN~\cite{wang2021real}  
&21.54 &0.5840 &0.2744 &0.5887 &\textbf{69.4185}  
&0.4898 &59.6803  
&0.6189 &\textbf{64.4957} \\
SwinIR~\cite{liang2021swinir} 
&21.53 &0.5901 &0.2652 &0.6096 &69.0808  
&0.4635 &59.6491  
&0.6050 &\underline{64.6984} \\
LDM-50~\cite{rombach2022high} 
&22.00 &0.5700 &0.2776 &0.5791 &66.1659  
&0.4907 &54.3910  
&0.5511 &55.8246 \\
ResShift-15~\cite{yue2024resshift}   
&\textbf{22.23} &0.5889 &\underline{0.2620} &\underline{0.6370} &68.7279  
&\underline{0.5998} &58.6835  
&\underline{0.6693} &62.8512 \\
% ResShift-4~\cite{yue2024efficient}   
% &\textbf{22.36} &\textbf{0.6005} &\textbf{0.2450} &\underline{0.6392} &67.6524  
% &0.5843 &56.0758  
% &0.6475 &60.8169 \\
QDM-L-15  
&\underline{22.16} &\textbf{0.5958} &\textbf{0.2452} &\textbf{0.6444} &69.1535 
&\textbf{0.6983} &\underline{61.0735}  
&\textbf{0.6918} &62.2867 \\
\bottomrule
\end{tabular}
}
\vspace{-0.2cm}
\label{table:merged}
\end{table*}

\subsection{Experimental Setup}
\noindent \textbf{Datasets.}
We train our proposed model on  ×4 real-world super-resolution (SR) tasks , and ×4 and x8 medical image SR tasks. For the real-world SR task, the training data integrates six established benchmarks: LSDIR~\citep{li2023lsdir}, DIV2K~\citep{agustsson2017ntire}, DIV8K~\citep{9021973}, OutdoorSceneTraining~\citep{wang2018recovering}, Flickr2K~\citep{Lim_2017_CVPR_Workshops}, and a subset of 10,000 facial images from the FFHQ dataset~\citep{karras2017progressive}.

For the medical CT SR task, we employ clinical CT scans from two well-established segmentation challenges: HaN-Seg~\citep{podobnik2023han} (focused on head-and-neck organ segmentation) and SegRap2023~\citep{luo2025segrap2023} (targeting naso-pharyngeal cancer radiotherapy planning). All medical images are partitioned into training and test sets using a 19:1 ratio to ensure rigorous evaluation.

To comprehensively assess real-world performance, we utilize two dedicated test benchmarks:
\textit{RealSR}~\citep{cai2019toward} which contains 100 authentic LR-HR pairs captured using Canon 5D3 and Nikon D810 DSLRs, and  \textit{RealSet80}~\citep{yue2024resshift} which curates 80 challenging low-resolution images from diverse internet sources and existing datasets. We also construct \textit{LSDIR-Test} benchmark using center-cropped $512\!\times\!512$ images from the LSDIR test set~\citep{li2023lsdir}. For the medical CT SR task, we create two benchmarks: \textit{Med-SR4} ($\times 4$ super-resolution task) and \textit{Med-SR8} ($\times 8$ super-resolution task). More details about data preprocessing, evaluation metrics as well as training process are in Appendix~\ref{appendix: implementation details}.

\begin{table}[t]
\centering
\small % Reduce font size to fit within one column
\caption{Quantitative comparisons of various methods on \textit{Med-SR4} and \textit{Med-SR8}. The best and second-best results are \textbf{bold} and \underline{underlined}, respectively.}
\setlength{\tabcolsep}{4pt} % Reduce horizontal padding
\renewcommand{\arraystretch}{1.1} % Adjust row height for better readability
% \vspace{-0.2cm}
\resizebox{\columnwidth}{!}{
\begin{tabular}{clcccc}
\toprule
\multirow{2}{*}{\textbf{Benchmarks}}& \multirow{2}{*}{\textbf{Methods}} & \multicolumn{3}{c}{\textbf{Metrics}} &\multirow{2}{*}{\textbf{FLOPs(T)}} \\ 
\cmidrule{3-5}
&  & \textbf{PSNR~$\uparrow$} & \textbf{SSIM~$\uparrow$} & \textbf{LPIPS$~\downarrow$} \\
\midrule
\multirow{5}{*}{\makecell[c]{\textit{Med-SR4}\\(128→512)}}
&LDM-50~\cite{rombach2022high}  &34.09 &0.9751 &\underline{0.0111}  &21.78\\
&ResShift-15~\cite{yue2024resshift} &\textbf{35.84} &\textbf{0.9828} &\textbf{0.0097} &6.54\\
&QDM-S-15 &33.99 &0.9766 &0.0162 &0.74\\
&QDM-B-15 &35.10 & 0.9804 &0.0147 &2.84\\
&QDM-L-15 &\underline{35.75} &\underline{0.9824} &0.0120 &10.00\\
\midrule
\multirow{5}{*}{\makecell[c]{\textit{Med-SR8}\\(64→512)}}
&LDM-50~\cite{rombach2022high} &31.00 &0.9635 &0.0162 &21.78\\
&ResShift-15~\cite{yue2024resshift} &\underline{32.63} &\underline{0.9728} &\textbf{0.0156} &6.54\\
&QDM-S-15 &30.39 &0.9617 &0.0240 &0.68\\
&QDM-B-15 &31.69 &0.9681 &0.0198 &2.58\\
&QDM-L-15 &\textbf{33.05} &\textbf{0.9743} &\underline{0.0159} &9.08\\
\bottomrule
\end{tabular}
}
\vspace{-0.2cm}
\label{table:medical results}
\end{table}

\subsection{Model Analysis}
\label{subsec:model analysis}
We benchmark our method against contemporary super-resolution approaches spanning two dominant paradigms: (1) \textit{Pretrained-Prior-Based Methods}, including StableSR~\cite{wang2024exploiting}, DiffBIR~\cite{lin2024diffbir}, SeeSR~\cite{wu2024seesr}, OSEDiff~\cite{wu2025one}, InvSR~\cite{yue2024arbitrary} and SinSR~\cite{wang2024sinsr}, which inherit generative priors from Stable Diffusion~\cite{rombach2022high, podell2023sdxl} or employ distillation from a pretrained model; (2) \textit{Training-from-scratch Methods}, comprising BSRGAN~\cite{zhang2021designing}, RealESRGAN~\cite{wang2021real}, and SwinIR~\cite{liang2021swinir}, LDM~\cite{rombach2022high}, and ResShift~\cite{yue2024resshift}.  For sampling consistency, we configure 50 diffusion steps for StableSR, DiffBIR, SeeSR, and LDM. Method-specific configurations are preserved for OSEDiff (1 steps), InvSR (1 steps), SinSR (1 step), and ResShift (15 steps) following their official implementations. 

\begin{table}[t]
\centering
\caption{Tumor Region Reconstruction Results (64$\rightarrow$512)}
% \vspace{-0.3cm}
\label{tab:roi_metrics}
\resizebox{\columnwidth}{!}{%
\begin{tabular}{@{}llccc@{}}
\toprule
Dataset & Metric & LDM & ResShift & QDM-L \\
\midrule
\multirow{2}{*}{SegRap CT} & PSNR$\uparrow$ & 18.51 & 20.97 & \textbf{22.71} \\
& SSIM$\uparrow$ & 0.5961 & 0.7237 & \textbf{0.7977} \\
\midrule
\multirow{2}{*}{\begin{tabular}[c]{@{}l@{}}SegRap CECT \\ (Out-of-Distribution)\end{tabular}} & PSNR$\uparrow$ & 17.46 & 19.27 & \textbf{20.31} \\
& SSIM$\uparrow$ & 0.5388 & 0.6449 & \textbf{0.6947} \\
\bottomrule
\end{tabular}
}
% \vspace{-0.5cm}
\end{table}

As demonstrated by the qualitative and quantitative results in Fig.~\ref{fig:comparison}, Table~\ref{table:merged}, and Table~\ref{table:medical results}, our proposed model outperforms or achieves comparable performance to state-of-the-art training-from-scratch methods across multiple benchmarks. While it still falls short of prior-based methods, it is worth emphasizing that such methods face significant practical limitations in medical imaging applications due to the absence of foundational generative models (e.g., Stable Diffusion) tailored to this domain. Furthermore, Table~\ref{table:medical results} highlights the computational efficiency of our QDM-B/S model: it delivers competitive performance relative to LDM and ResShift while requiring substantially less computational overhead. In contrast, lightweight methods like BSRGAN and SwinIR, though computationally efficient, achieve markedly inferior results. More importantly, Table~\ref{tab:roi_metrics} demonstrates that QDM attains superior PSNR and SSIM scores in tumor regions on both SegRap Tumor CTs and the more challenging out-of-distribution SegRap Tumor CECTs, outperforming LDM and ResShift. This underscores QDM’s effectiveness in enhancing clinically critical pathological areas. Additional qualitative results are provided in Appendix~\ref{appendix:qualitative results}.

\subsection{Ablation Studies}
\noindent \textbf{Quadtree mask threshold.}
As detailed in Section~\ref{subsec:model architecture}, the quadtree mask density controls computation allocation by adaptively partitioning images into homogeneous regions (processed via efficient upstream) and detail-rich areas (handled by high-resolution downstream). Table~\ref{table:ablation quadtree threshold} demonstrates this on \textit{Med-SR8}: At $s=0$ (lossless mode), QDM-L matches full-mask performance while cutting computation by $40.3\%$, confirming that non-uniform prediction suffices for medical SR. Increasing $s$ to $0.90$ reduces computation to $15.5\%$ baseline ($6.5\times$ less) with only $0.34$ dB PSNR drop, favoring efficiency-critical clinical use.
To balance performance and efficiency, we set the threshold $s=0.15$ for all medical SR experiments.

\begin{table}[t]
\centering
\small
\caption{Ablation study on threshold $s$: quadtree mask density and computational cost in QDM-L for \textit{Med-8}.}
\setlength{\tabcolsep}{3.5pt}  
\renewcommand{\arraystretch}{1.05}
% \vspace{-0.2cm}
\resizebox{\columnwidth}{!}{
\begin{tabular}{@{}lccccc@{}}
\toprule
\textbf{Threshold} & \textbf{PSNR}~$\uparrow$ & \textbf{SSIM}~$\uparrow$ & \textbf{LPIPS}~$\downarrow$ &\textbf{\%Density} &\textbf{FLOPs(T)}\\
\midrule
 Full Mask  & 33.06 & 0.9744 &0.0144 & 100.00 & 36.64\\
 0.00  & 33.06 & 0.9744 &0.0144 &49.14 &21.88\\
  0.15  & 33.05 & 0.9743 & 0.0159 &9.59 &9.08\\
 0.30  & 33.01 & 0.9739 & 0.0174 &7.64 &8.18\\
 0.60  & 32.85 & 0.9728 & 0.0201 &4.56 & 7.30\\
 0.90  & 32.67 & 0.9719 & 0.0221 &2.25  & 5.68\\
\bottomrule
\end{tabular}
}
\label{table:ablation quadtree threshold}
% \vspace{-0.3cm}
\end{table}

\noindent \textbf{Upstream patch size~\& Downstream window size.}
As shown in Table~\ref{table:patch and grid size ablation}, we ablate upstream patch and downstream window sizes. Smaller upstream patches enhance fine-grained details (benefiting both streams) but increase self/cross-attention costs. Larger downstream windows expand local context, improving performance at greater computational expense. To balance performance and efficiency, we set the upstream patch size and downstream window size to 8 for both real-world and medical SR experiments.

\begin{table}[t]
\centering
\small
\caption{Ablation study on upstream patch and downstream window sizes for QDM-B on \textit{Med-4} and \textit{Med-8} (threshold s=0.00).}
\setlength{\tabcolsep}{3.5pt}  
\renewcommand{\arraystretch}{1.05}
% \vspace{-0.2cm}
\resizebox{\columnwidth}{!}{
\begin{tabular}{@{}clcccc@{}}
\toprule
\textbf{Benchmark} & \textbf{Param} & \textbf{PSNR}~$\uparrow$ & \textbf{SSIM}~$\uparrow$ & \textbf{LPIPS}~$\downarrow$ &\textbf{FLOPs(T)} \\
\midrule
\multirow{5}{*}{\makecell[c]{\textit{Med-SR4}\\(128→512)}} 
& Patch (4)  &35.18 & 0.9806 &0.0113 &8.20\\
& Patch (8)  & 35.10 & 0.9804 & 0.0115 &6.28\\

\cmidrule{2-6}
& Window (4)  & 34.91  & 0.9797 & 0.0119  &5.98\\
& Window (8)   & 35.10 & 0.9804 & 0.0115 &6.28\\
& Window (16)  & 35.15 & 0.9807 & 0.0114  &7.16   \\
\midrule
\multirow{5}{*}{\makecell[c]{\textit{Med-SR8}\\(64→512)}}
& Patch (4)  & 31.88   & 0.9690     & 0.0172     &8.14 \\
& Patch (8)  & 31.69   & 0.9682     & 0.0178     &6.22\\
\cmidrule{2-6}
& Window (4)   & 31.55 & 0.9673 & 0.0181   &5.88  \\
& Window (8)   & 31.69 & 0.9682 & 0.0178   &6.22\\
& Window (16)  & 31.73 & 0.9683 & 0.0175   &7.12  \\
\bottomrule
\end{tabular}
}
\label{table:patch and grid size ablation}
% \vspace{-0.3cm}
\end{table}

\section{Conclusion}
\label{sec:conclusion}
In this work, we present QDM, a novel diffusion-based super-resolution (SR) method that leverages quadtree-based region-adaptive sparse diffusion to enhance detail-rich image regions while minimizing computational overhead in homogeneous areas. 
The proposed framework incorporates two key innovations: 1) a quadtree-based sparse diffusion model that rapidly detects detail-rich regions and only performs diffusion process in such areas, and 2) a two-stream architecture that decouples global and local feature extraction. The upstream branch captures broad contextual information using large patch sizes, while the downstream branch employs cross-attention to refine high-detail chunks identified via quadtree masks. This design achieves two advantages. First, it eliminates the need to store high-dimensional feature maps, significantly reducing memory overhead. Second, it dynamically minimizes computation by selectively processing only salient regions in the downstream branch. 
Extensive experiments on real-world SR and medical CT reconstruction tasks demonstrate that QDM outperforms existing training-from-scratch diffusion models in both reconstruction quality and computational efficiency. 
These results underscore the potential of region-adaptive diffusion strategies for resource-conscious high-fidelity image enhancement.

\clearpage
{
    \small
    \bibliographystyle{ieeenat_fullname}
    \bibliography{main}
}

\clearpage
\setcounter{page}{1}
\appendix
\section{Different Configurations of QDM}
\label{appendix:configurations}
We present the detailed configurations of our proposed QDM models as well as their number of  parameters in Table \ref{table:QDM configs}. 
\begin{table}[ht]
    \centering
    \caption{\textbf{Details of QDM models.} Our configurations for Small (S), Base (B) and Large (L) models are based on those of the DiT model \citep{peebles2023scalable}. To achieve a comparable parameter count, we halve the number of layers from the original DiT configurations.}
    \label{table:QDM configs}
    % \vspace{-0.2cm}
    \resizebox{\columnwidth}{!}{% Scale table to fit page width
    \begin{tabular}{lcccc}
        \toprule
         Model & Layers $N$ & Hidden size & Heads & \#Parameters(M)
        \\
        \midrule
        \textbf{QDM-S} &6 &384 &6 &50.87\\
        \textbf{QDM-B} &6 &768 &12 &199.06\\
        \textbf{QDM-L} &12 &1024 &16 &691.94\\
        \bottomrule
    \end{tabular}
    }
\end{table}

\begin{figure}[ht]
\centering
\includegraphics[width=\columnwidth]{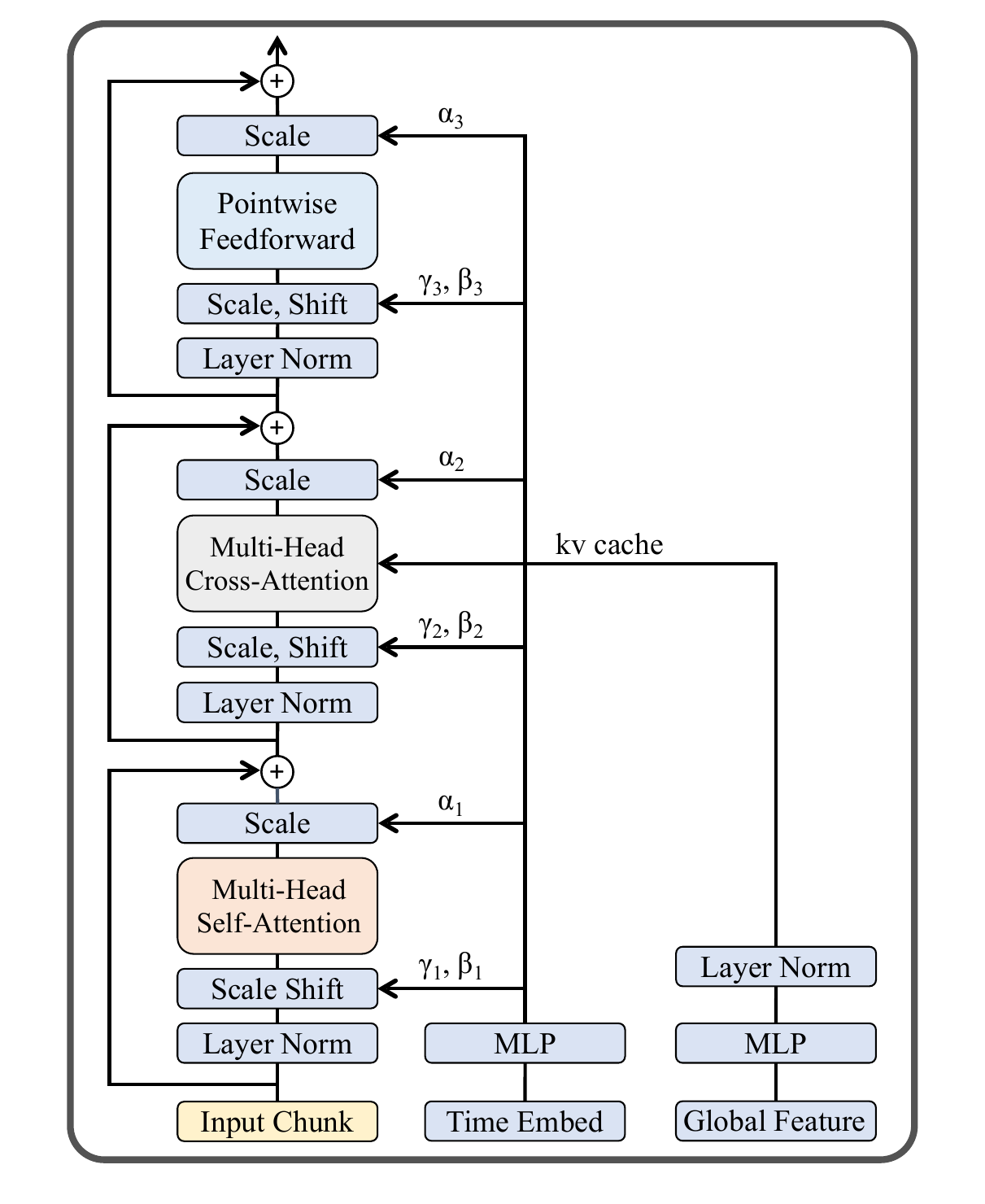}
\caption{\textbf{Architecture of the cross-attention block.} Each block sequentially processes features through: (1) a multi-head self-attention (MSA) layer for local intra-window interactions, (2) a multi-head cross-attention (MCA) layer to fuse global context from upstream features, and (3) a two-layer feedforward network with GELU activation. All outputs from MSA, MCA, and feedforward are combined with residual connections to their respective inputs. Time-step conditioning is applied via adaLN-Zero layers, while the upstream global features are pre-projected into reusable key-value (kv) pairs to minimize redundant computations.}  
\label{fig:cross attn block}
\vspace{-0.2cm}  
\end{figure}

\section{Implementation Details}
\label{appendix: implementation details}
\noindent \textbf{Data Processing.}
For the real-world SR task, we adopt fixed-size $512\!\times\!512$ image crops during training. Following recent work~\citep{yang2024inf}, training images exceeding this resolution undergo either: (1) direct random cropping, or (2) resizing the shorter edge to 512 pixels before random cropping. We synthesize LR/HR pairs using the degradation pipeline from Real-ESRGAN~\citep{wang2021real}. 

Beyond real-world test benchmarks, we construct \textit{LSDIR-Test} using center-cropped $512\!\times\!512$ images from the LSDIR test set~\citep{li2023lsdir}, processed through identical degradation parameters as the training data.

For the medical CT SR task, we use the first degradation stage of Real-ESRGAN pipeline~\citep{wang2021real} simulating clinical noise artifacts. Both training and test images are first downsampled to $512\!\times\!512$ resolution as HR references, then processed through the adopted pipeline to obtain the corresponding LR image. Thus, we can create two benchmarks: \textit{Med-SR4} ($\times 4$ super-resolution task) and \textit{Med-SR8} ($\times 8$ super-resolution task).

\noindent \textbf{Evaluation Metrics.} The performance of different methods was evaluated using both reference-based and non-reference metrics. For reference-based assessment, we employed PSNR, SSIM~\citep{wang2004image}, and LPIPS~\citep{zhang2018unreasonable}. To better align with human perceptual judgments in generative super-resolution, we also incorporated non-reference metrics: CLIPIQA~\citep{wang2023exploring} and MUSIQ~\citep{ke2021musiq}. All LPIPS, CLIPIQA, and MUSIQ results were computed following the official implementations provided in IQA-PyTorch~\citep{pyiqa}\footnote{\url{https://github.com/chaofengc/IQA-PyTorch}}.

For the \textit{LSDIR-Test} and \textit{RealSR} datasets, both reference and non-reference metrics were applied to ensure comprehensive evaluation. In contrast, only non-reference metrics were used for \textit{RealSet80} due to the lack of ground truth images. Notably, for real-world super-resolution tasks, PSNR and SSIM were computed in the luminance (Y) channel of the YCbCr color space, while other metrics were calculated directly in RGB space. Evaluations on the \textit{Med-SR4} and \textit{Med-SR8} datasets relied solely on reference metrics, as non-reference perceptual scores were deemed less relevant.

\noindent \textbf{Training Process \& Hyperparamters.} In real-world SR task, following prior work (LDM~\citep{rombach2022high}, ResShift~\citep{yue2024resshift}), our architecture operates in latent space using a Vector Quantized GAN (VQGAN, \citep{esser2021taming}) with a downsampling factor of 4. The model was trained for 150,000 iterations using the dual-stream training objective (Eq.~\ref{eq:dual-stream training objective}), with a batch size of 64 on eight NVIDIA A100 80GB PCIe GPUs. We employed the Adam optimizer with a learning rate of $5 \times 10^{-5}$, followed by a 50,000-iteration fine-tuning phase using the perceptual training objective (Eq.~\ref{eq:perceptual training objective}). For all real-world experiments, we set the quadtree threshold $s=0.00$.

For medical CT scan super-resolution, we first trained a KL-regularized VAE ~\citep{kingma2013auto,rombach2022high} on the medical dataset with a 4$\times$ downsampling factor. Our proposed model was then trained in this latent space for 50,000 iterations. To ensure fair comparison, ResShift and LDM were reimplemented and trained under identical conditions (50k iterations), while BSRGAN~\citep{zhang2021designing} and SwinIR~\citep{liang2021swinir} were trained for 100,000 iterations to account for their distinct optimization requirements. For all medical SR experiments, we set the quadtree threshold $s=0.15$. For both tasks, the upstream patch size and downstream window size were set to 8. The low-resolution (LR) image was concatenated to the latent feature map before being passed to QDM, with a maximum of 64 chunks processed in parallel. Loss weightings $\lambda_1$ and $\lambda_2$ were both set to 1, and $\lambda_3$ was set to 0.1.

Besides, Figure~\ref{fig:cross attn block} provides a detailed structure of cross-attention block presented in Figure~\ref{fig:framework}. Each cross-attention block processes features sequentially through three components: a local intra-window self-attention, a cross-attention mechanism that integrates global context from upstream features, and a feedforward network. It's important to note that the upstream global features are pre-projected into reusable key-value (kv) pairs to minimize redundant computations. The MLP ratio is set to the default value of 4.

\noindent \textbf{Ultra-High-Resolution Inference.} To handle ultra-high-resolution (UHR) images beyond GPU memory limits, we adopt a patch-based inference strategy with Gaussian-weighted fusion proposed in \cite{yue2024arbitrary}. The input image $I_{\mathrm{LR}}\in\mathbb{R}^{B\times C\times H\times W}$ is divided into overlapping patches of size $P\times P$ and stride $S\le P$. Each patch is super-resolved individually to $\hat{p}_k\in\mathbb{R}^{C\times (s_fP)\times (s_fP)}$ and accumulated into a global canvas with corresponding Gaussian weights $w_k(x)$ generated by separable 1D kernels. The final output is obtained by normalized fusion:
\[
I_{\mathrm{HR}}(x)=\frac{\sum_k w_k(x)\,\hat{p}_k(x)}{\sum_k w_k(x)}.
\]
This Gaussian overlap-add formulation ensures seamless blending between neighboring patches while enabling scalable processing of arbitrarily large images. In practice, we set $P\!=128$, $S\!=\!112$, and accumulate in \texttt{float64} precision for numerical stability.

\section{Additional Experimental Results}
\label{appendix: additional experimental results}

\begin{table}[ht]
\centering
\caption{Downstream-only and prior-based methods on Med-8}
\vspace{-0.2cm} 
\label{tab:downstream-prior}
\resizebox{0.9\columnwidth}{!}{%
\begin{tabular}{@{}lcccc@{}}
\toprule
\multirow{2}{*}{Metric} & \multirow{2}{*}{Downstream-only} & \multicolumn{2}{c}{Prior-based} & \multirow{2}{*}{QDM-L}\\
\cmidrule(lr){3-4}
    &   & DiffBIR      & SeeSR\\
\midrule
PSNR$\uparrow$  &31.75  & 23.62        & 26.87        & \textbf{33.05}           \\
SSIM$\uparrow$  &0.9685  & 0.6553       & 0.6665       & \textbf{0.9743}          \\
LPIPS$\downarrow$ &0.0181  & 0.3686       & 0.0686       & \textbf{0.0159}          \\
\bottomrule
\end{tabular}%
}
\end{table}

\begin{table}[ht]
\centering
\caption{Quantitative comparisons of the proposed QDM-L equipped with different thresholds, on the
benchmark \textit{LSDIR-Test}.}
\vspace{-0.2cm}
\resizebox{\columnwidth}{!}{% Scale table to fit page width
\begin{tabular}{lccccccc}
\toprule
\multirow{2}{*}{\textbf{Threshold}} &\multirow{2}{*}{\textbf{Density}}
&\multirow{2}{*}{\textbf{MACs(T)}}
& \multicolumn{3}{c}{\textbf{Reference Metrics}} & \multicolumn{2}{c}{\textbf{Non-Reference Metrics}} \\
\cmidrule(lr){4-6} \cmidrule(lr){7-8}
& & & \textbf{PSNR} $\uparrow$ & \textbf{SSIM} $\uparrow$ & \textbf{LPIPS} $\downarrow$  & \textbf{CLIPIQA$\uparrow$} & \textbf{MUSIQ$\uparrow$} \\
\midrule
Full Mask &100.00 & 18.32 &22.16 &0.5958 &0.2452 &0.6444 &69.1535\\
$s=0.00$ & 99.41 & 18.32 &22.16 &0.5958 &0.2452 &0.6444 &69.1535\\
$s=0.03$ &90.42 &18.19  &22.34 &0.6059 &0.2658 &0.5646 &68.4705\\
$s=0.06$ &80.18 & 17.98 &22.50 &0.6100 &0.2997 &0.5016 &65.9053\\
$s=0.09$ &70.37 &17.73  &22.62 &0.6105 &0.3340 &0.4493 &62.3469\\
\bottomrule
\end{tabular}
}
\label{table:real-world ablation quadtree threshold}
\end{table}

\noindent \textbf{Downstream-only Design on \textit{Med-SR8}.} We designed the dual‐stream architecture to leverage the complementary strengths of both branches. The downstream branch focuses on local, detail-rich regions while the upstream branch captures global context and applies minimum refinement on homogeneous regions. Here, we conducted experiments to validate that without the upstream branch, the model cannot capture the global view and is prone to overlook large‐scale structure, leading to degraded performance, as shown in Table~\ref{tab:downstream-prior}. More importantly, a downstream‐only design lacks the ability to allocate computation adaptively across different regions of the image.

\noindent \textbf{Pretrained-Prior-Based Methods on \textit{Med-SR8}.} We've included detailed results on natural images comparing to several prior-based methods in Table~\ref{table:merged}. To test their generalization abilities, we evaluated two methods (DiffBIR and SeeSR) on the Med-8 benchmark. Neither method yielded satisfactory reconstructions in the medical domain, with notably lower performance compared to QDM (Table~\ref{tab:downstream-prior}). These results highlight a substantial generalization gap, suggesting that such methods may require significant domain-specific data and fine-tuning.

\noindent \textbf{Ablation study of quadtree threshold on \textit{LSDIR-Test}.}
We also conducted an ablation study on the quadtree threshold using real-world benchmarks. However, real-world images often contain significant noise and lack large homogeneous regions, making them less suited for adaptive computation. As shown in Table~\ref{table:ablation quadtree threshold}, at s=0 (lossless mode), QDM-L achieves performance identical to the full-mask baseline but exhibits minimal computation reduction, as the mask remains nearly dense. Increasing $s$ to $0.09$ reduces computation by $3.22\%$ compared to the baseline, but this comes at a significant performance cost. This suggests that higher quadtree thresholds are not well-suited for super-resolution tasks involving noisy and complex low-resolution inputs.

\section{Qualitative comparisons of different methods.} 
\label{appendix:qualitative results}
\begin{figure*}[ht]
\centering
\includegraphics[width=\textwidth]{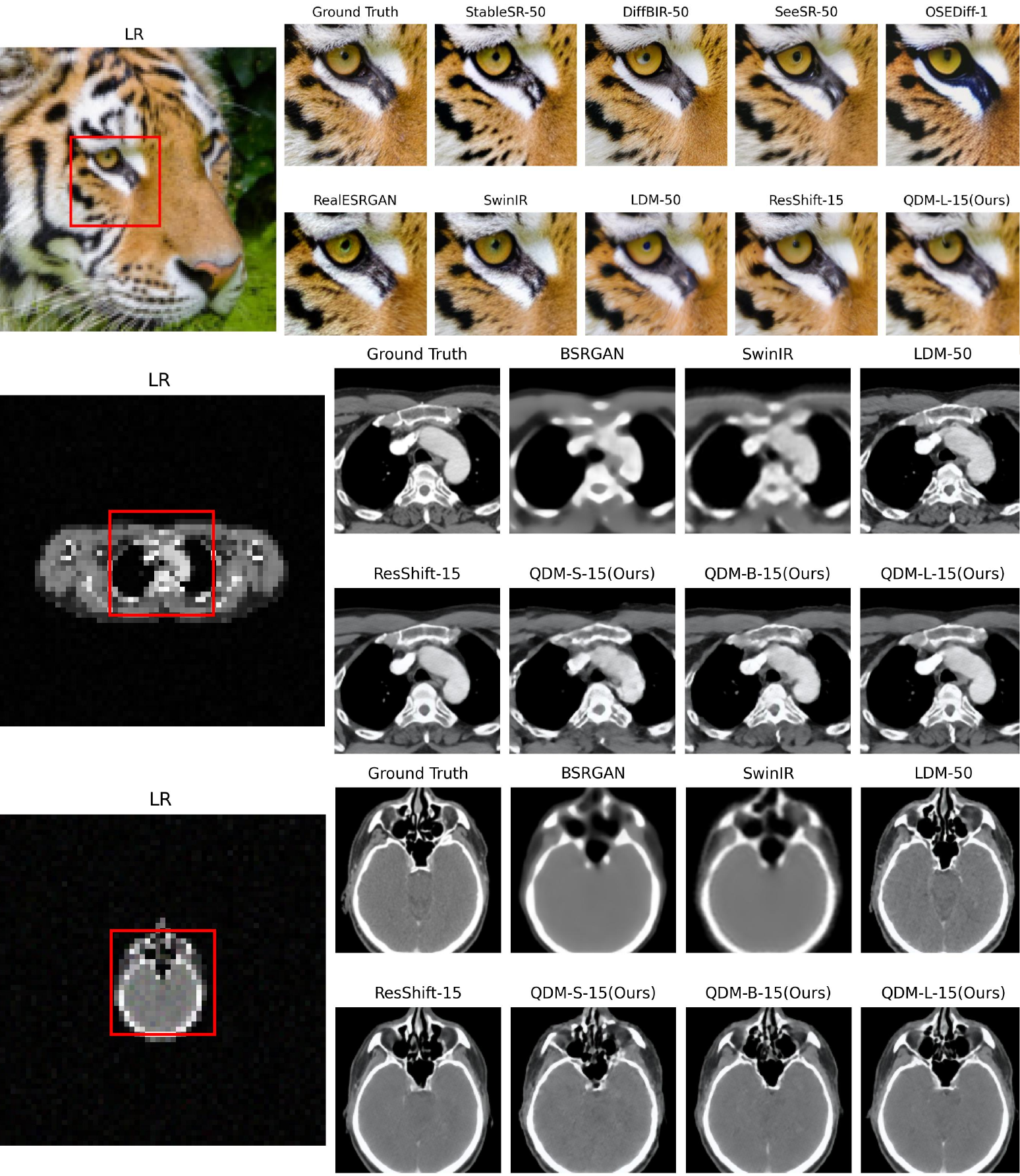}
\vspace{-0.5cm}
\caption{Visual comparison of different methods on real-world images and medical CT datasets. Zoom in for finer details. }  
\label{fig:comparison appendix 1}
\vspace{-0.5cm}  
\end{figure*}

\begin{figure*}[ht]
\centering
\includegraphics[width=\textwidth]{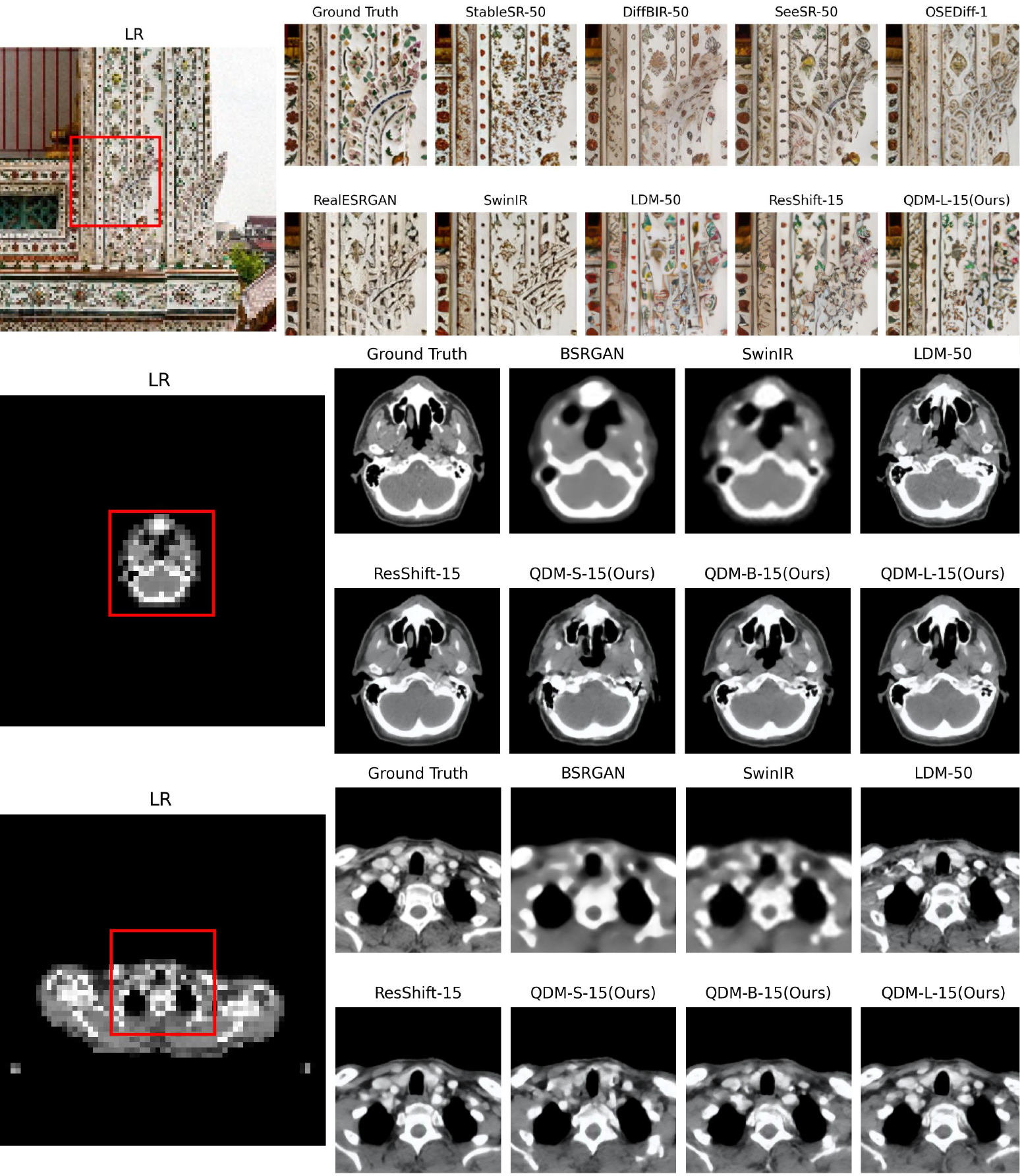}
\vspace{-0.5cm}
\caption{Visual comparison of different methods on real-world images and medical CT datasets. Zoom in for finer details. }  
\label{fig:comparison appendix 2}
\vspace{-0.5cm}  
\end{figure*}

\begin{figure*}[t]  % h=here, t=top, b=bottom, p=page
    \centering
    \includegraphics[width=0.6\textwidth]{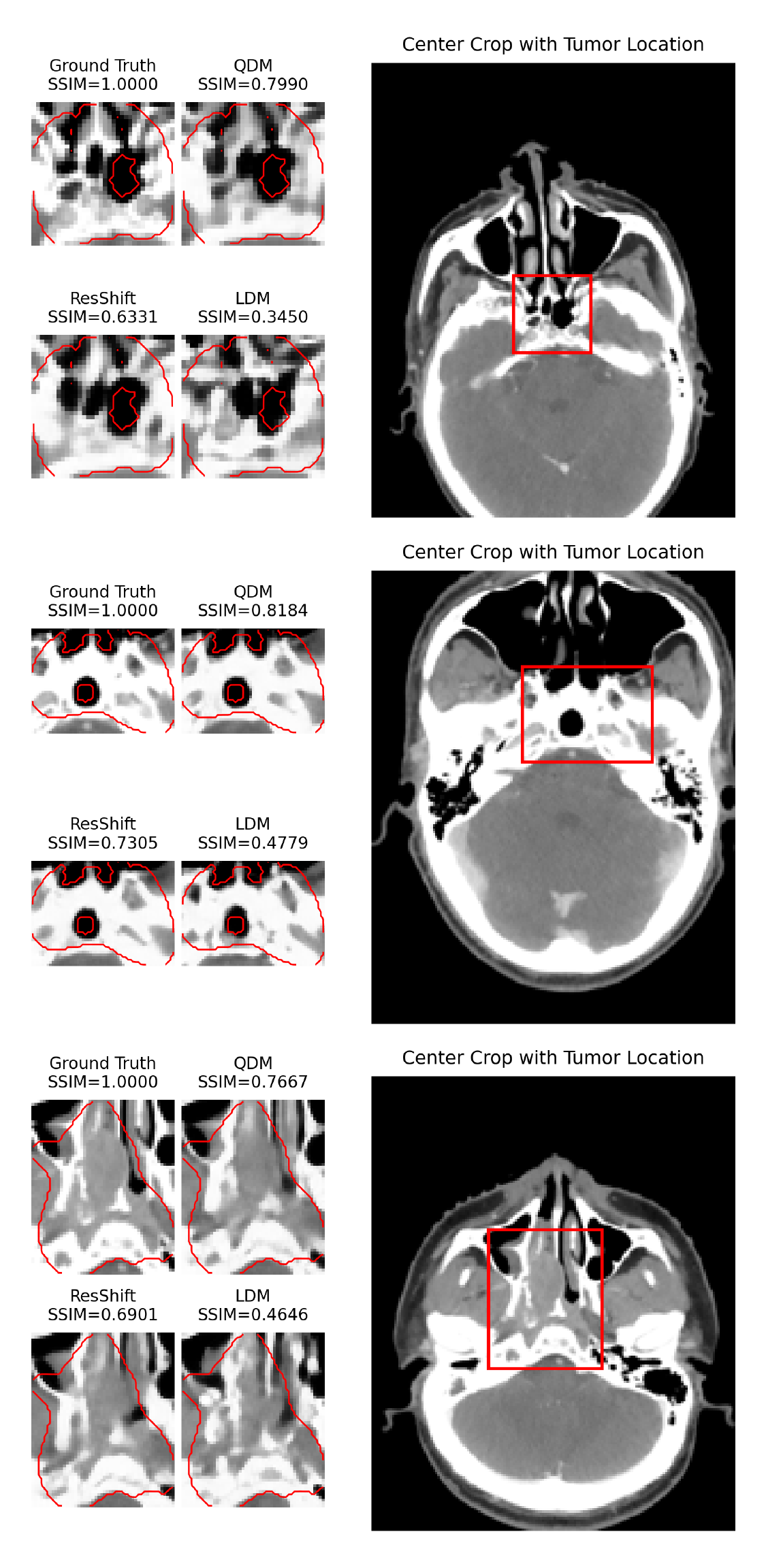}
    \vspace{-0.5cm}
    \caption{Qualitative comparison of CECTs in the tumor region.}
    \label{fig:ROI_comparisons}
\end{figure*}

We also include more qualitative comparisons of different methods in Figure~\ref{fig:comparison appendix 1}, Figure~\ref{fig:comparison appendix 2} and Figure~\ref{fig:ROI_comparisons}.

\vspace{-1em}

\end{document}